\documentclass[conference]{IEEEtran}
\IEEEoverridecommandlockouts

\usepackage{cite}
\usepackage{amsmath,amssymb,amsfonts}
\usepackage{algorithmic}
\usepackage{textcomp}
\usepackage{graphicx}
\usepackage{mathptmx}
\usepackage{times}
\usepackage{hyperref}
\usepackage{booktabs}
\usepackage{multirow}
\usepackage{tablefootnote}

\def\BibTeX{{\rm B\kern-.05em{\sc i\kern-.025em b}\kern-.08em
    T\kern-.1667em\lower.7ex\hbox{E}\kern-.125emX}}

\renewcommand{\sec}[1]{Section~\ref{#1}}
\newcommand{\fig}[1]{Figure~\ref{#1}}

\newcommand{\tab}[1]{Table~\ref{#1}}

\begin{document}
\title{Deep Reinforcement Learning for the Joint Control of Traffic Light Signaling and Vehicle Speed Advice}

\author{%
\IEEEauthorblockN{%
    Johannes V. S. Busch\IEEEauthorrefmark{1}\IEEEauthorrefmark{3}, %
    Robert Voelckner\IEEEauthorrefmark{2}, %
    Peter Sossalla\IEEEauthorrefmark{2}\IEEEauthorrefmark{4}, %
    Christian L. Vielhaus\IEEEauthorrefmark{2},\\%
    Roberto Calandra\IEEEauthorrefmark{1}\IEEEauthorrefmark{3}, and %
    Frank H. P. Fitzek\IEEEauthorrefmark{2}\IEEEauthorrefmark{3}%
}
\IEEEauthorblockA{\IEEEauthorrefmark{1}Learning, Adaptive Systems, and Robotics (LASR) Lab, TU Dresden, Germany}%
\IEEEauthorblockA{\IEEEauthorrefmark{2}Deutsche Telekom Chair of Communication Networks, TU Dresden, Germany}%
\IEEEauthorblockA{\IEEEauthorrefmark{3}Center for Tactile Internet with Human-in-the-Loop (CeTI), TU Dresden, Germany}%
\IEEEauthorblockA{\IEEEauthorrefmark{4}Audi AG, Ingolstadt, Germany}%
\thanks{Funded by the German Research Foundation (DFG, Deutsche Forschungsgemeinschaft) as part of Germany’s Excellence Strategy – EXC 2050/1 – Project ID 390696704 – Cluster of Excellence “Centre for Tactile Internet with Human-in-the-Loop” (CeTI) of Technische Universität Dresden.}%
\thanks{Corresponding author: jbusch@lasr.org}%
\thanks{Code Available Online: \url{https://github.com/robvoe/RELESAS}}%
}

\maketitle

\begin{abstract}
    Traffic congestion in dense urban centers presents an economical and environmental burden.
    In recent years, the availability of vehicle-to-anything communication allows for the transmission of detailed vehicle states to the infrastructure that can be used for intelligent traffic light control.
    The other way around, the infrastructure can provide vehicles with advice on driving behavior, such as appropriate velocities, which can improve the efficacy of the traffic system.
    Several research works applied deep reinforcement learning to either traffic light control or vehicle speed advice.
    In this work, we propose a first attempt to jointly learn the control of both.
    We show this to improve the efficacy of traffic systems.
    In our experiments, the joint control approach reduces average vehicle trip delays, w.r.t. controlling only traffic lights, in eight out of eleven benchmark scenarios.
    Analyzing the qualitative behavior of the vehicle speed advice policy, we observe that this is achieved by smoothing out the velocity profile of vehicles nearby a traffic light.
    Learning joint control of traffic signaling and speed advice in the real world could help to reduce congestion and mitigate the economical and environmental repercussions of today's traffic systems.
\end{abstract}

\section{Introduction}
    Traffic congestion is a major source of delay in transportation networks and results in significant economic and environmental repercussions.
    Minimizing the delay caused in traffic systems is thus one of the primary concerns of traffic research.
    Due to spatial constraints, it is often difficult to increase road capacities by simply building wider roads.
    This is especially true in dense urban centers, where the problem of traffic congestion is particularly severe.
    It is therefore important to increase the capacity of existing traffic systems through the intelligent allocation of given resources.
    
    A particularly important measure for traffic control are traffic lights.
    Past literature focused on the optimization of traffic light signaling to mitigate congestion \cite{Robertson1969TRANSYTControl, Robertson1991OptimizingMethod}.
    The increasing availability of vehicle-to-anything~(V2X) communication technologies, such as IEEE 802.11p or 5G-V2X~\cite{Anwar2019Physical802.11p}, allows the collection of a detailed traffic state that, in theory, could enable more informed traffic control decisions.
    Since most traditional methods are not designed to handle detailed state data, in recent years many researchers have successfully applied deep reinforcement learning~(DRL) algorithms to infer good traffic light control polices from data~\cite{Richter2006NaturalOptimisation, Zhang2021UsingControl, Busch2020OptimisedKnowledge}. 
    Many works implement complex multi-agent algorithms to learn intricate cooperation of traffic lights, while scaling to large scenarios~\cite{VanderPol2016CoordinatedControl, Wei2019CoLight:Control, Ge2019CooperativeControl, Chen2020TowardControl, Chu2020Multi-AgentControl}.
    Most publications claim to outperform previous state-of-the-art methods when evaluated on their own simulation scenarios. 
    However, in a set of established benchmark scenarios, Ault, et. al. were unable to reproduce these results~\cite{Ault2021ReinforcementControl}.
    Indeed, evaluated on said benchmark, the algorithms that performed best were relatively simple independent learners that implement no explicit cooperation mechanisms.

    \begin{figure}
        \centering
        \includegraphics[width=0.96\columnwidth]{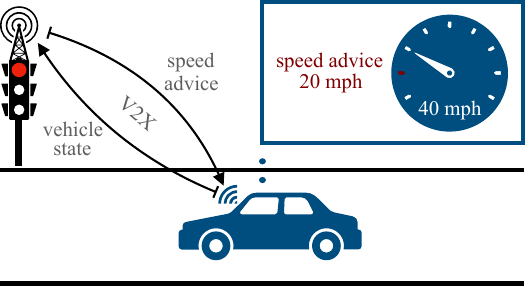}
        \caption{%
        The envisioned traffic system.
        Via V2X communication, vehicles share their current state so that the infrastructure can make more informed decisions.
        The infrastructure transmits advice on optimal velocities to vehicles that can be displayed on the dashboard.}
        \label{fig:concept}
    \end{figure}
    
    Another measure to control traffic systems, which has started to emerge, is vehicle speed advice.
    In contrast to speed limits, speed advice is not legally binding, but can be used to encourage more foresighted driving behavior that increases the safety and efficacy of traffic systems.
    It can be presented to drivers via adaptive street signs or can be made available on vehicles' dashboards via V2X communication.
    DRL has been used to control vehicle speed by adapting speed limits~\cite{Kusic2021DynamicLearning} or directly optimizing velocities of individual vehicles~\cite{Jiang2021DampenApproach, Yen2022DeepOptimization}. 
    
    In previous works, DRL has been applied to both traffic light control and vehicle speed advice individually.
    In this work, we will develop and test a first approach to jointly optimize the two measures.
    To do so, we build on the results of Ault, et. al.~\cite{Ault2021ReinforcementControl} by implementing DRL agents, as independent learners, that control both traffic light signaling and vehicle speed advice of individual vehicles.
    \fig{fig:concept} shows the envisioned system.
    We evaluate this approach both qualitatively, on an atomic single intersection scenario, as well as quantitatively on the benchmark scenarios from~\cite{Ault2021ReinforcementControl}.
    In our experiments, learning vehicle speed advice on top of traffic light signaling shows to improve the efficiency of traffic systems in terms of travel time. This improvement is achieved by smoothing out velocity profiles of vehicles that are approaching the intersection.
    This work presents a first attempt towards jointly optimizing traffic light signaling and vehicle speed advice.
    We expect future work to build on our results by exploring more evolved approaches of integrating the two kinds of agents.
    
\section{Background and Related Work}
    Reinforcement learning~(RL) is a subfield of machine learning in that an agent learns to sequentially make decisions, based on its current state, to maximize a numerical reward over time~\cite{Sutton2018ReinforcementIntroduction}.
    It learns through trial and error, receiving positive reinforcement for actions that lead to desired outcomes and negative reinforcement for actions that do not.
    This makes it a useful approach when it is difficult to derive an optimal policy from first principles.
    Many traditional RL approaches rely on tabular representations, limiting state and action spaces to be discrete~\cite{Sutton2018ReinforcementIntroduction}.
    Recent advances in RL have been driven by the application of Deep Neural Networks~(DNNs) as function approximators, which allow the application to complex problem domains with continuous state and action spaces.
    The combination of RL and DNNs is referred to as Deep Reinforcement Learning~(DRL).
    In this work we will use the Proximal Policy Optimization~(PPO) algorithm~\cite{Schulman2017ProximalAlgorithms}, which shows good convergence properties, allows for discrete and continuous action spaces, and is a popular algorithm in DRL. 
    In many real-world settings, multiple learning agents act in a shared environment, giving rise to certain multi-agent pathologies that can aggravate learning.
    Multi-agent RL~(MARL) methods implement mechanisms to deal with these issues.
    In this work we will follow the naive approach of independent learners, that simply ignores all multi-agent pathologies and implements all agents using single-agent methods~\cite{Hernandez-Leal2019ALearning}.

    Vehicle-to-anything~(V2X) communication enables the exchange of information between all entities of a traffic system, giving rise to a plethora off new application possibilities.
    Existing and emerging technologies, such as IEEE 802.11p or 5G-V2X~\cite{Anwar2019Physical802.11p}, promise reliable low latency communication over the air that matches the high safety requirements and tight real-time constraints of traffic systems.

    Traffic light control is a critical aspect of transportation infrastructure as it helps to ensure the safety and efficiency of vehicle and pedestrian movement in urban areas.
    Approaches to traffic light control can be categorized into fixed-time and adaptive control.
    Fixed-time control methods, like TRANSYT~\cite{Robertson1969TRANSYTControl}, use predetermined signal timing patterns based on expected traffic and road layout, not taking into account the present traffic condition.
    Adaptive control, like SCOOT~\cite{Robertson1991OptimizingMethod}, uses real-time data to continually adjust signal timing, based on the current traffic situation, with the goal of maximizing throughput and minimizing delays.
    In recent years, there has been increasing interest in the use of machine learning, and in particular DRL, to improve traffic light control~\cite{Wei2021RecentEvaluation}.
    Publications strongly differ in their implemented state space, which is ultimately defined by the sensing capabilities of the traffic infrastructure.
    \cite{Richter2006NaturalOptimisation} assume traffic information to originate from inductive loop sensors.
    Many works use more detailed traffic state information that could only be inferred from additional sensors, like traffic cameras, or through the communication of individual vehicle states via V2X communication.
    A popular representation, often associated with the availability of traffic cameras, is obtained by dividing roads into a high resolution grid, where every cell encodes spatial information like occupancy, speed, and acceleration~\cite{VanderPol2016CoordinatedControl, Ge2019CooperativeControl}.
    Other approaches, typically considered to obtain real-time information through V2X communication, use concatenated vehicle states to encode detailed traffic state information, such as the distance of the leading vehicle to the next intersection for every lane~\cite{Zhang2021UsingControl}, or position and velocity of the closest N vehicles~\cite{Busch2020OptimisedKnowledge}.
    Other distinguishing factors include the choice of action space, reward function, the choice of RL algorithm, and the cooperation of different intersections (for a detailed evaluation see~\cite{Wei2021RecentEvaluation}).
    Many publication claim to beat the former state of the art in their own simulation environment, mostly by implementing sophisticated methods from MARL~\cite{VanderPol2016CoordinatedControl,Ge2019CooperativeControl,Wei2019CoLight:Control,Chu2020Multi-AgentControl,Chen2020TowardControl}.
    However, when comparing algorithms on a broad benchmark of traffic scenarios, Ault, et. al. show that learning with independent learners, without intricate cooperation strategies or sharing of DNN parameters, outperforms other approaches~\cite{Ault2021ReinforcementControl}.
    In this paper, we heavily leverage the results obtained by Ault, et. al. in adopting their methodology and building our own methods on top of it, as well as using their set of benchmark scenarios.
    A detailed explanation of our modifications can be found in \sec{sec:methodology}; a brief overview of the benchmarks and metrics introduced in~\cite{Ault2021ReinforcementControl}, in \sec{sec:exp_setup}.

    Adaptive speed control are measures to adapt speed limits to the current traffic situation.
    One option that has been deployed for several decades are adaptive roadside speed signs that can lower speed limits in case of congested traffic.
    This has shown to reduce accident rates and increase traffic capacities~\cite{Weikl2013TrafficAutobahn}.
    Lately, there has been increasing interest in the use of advanced technologies, such as DRL, to improve speed control systems.
    In~\cite{Kusic2021DynamicLearning}, DRL is used to control adaptive speed signs along a road to optimize vehicle delays.
    Other approaches directly control the acceleration (and therefore the speed) of individual vehicles to dampen oscillations in dense traffic~\cite{Jiang2021DampenApproach, Yen2022DeepOptimization}.
    Vehicle speed advice is a softer, not legally binding alternative to speed limits that can be used to advise drivers on appropriate driving velocities.
    Speed advice can be communicated to drivers through adaptive street signs.
    More recently, car manufacturers have implemented mechanisms to display speed advice, transmitted via V2X communication, inside the vehicle~\cite{Stahlmann2018ExploringResults}.
    The application of speed advice shows to improve traffic flow and reduce CO$_2$ emissions~\cite{Eckhoff2013PotentialsSystems}.
    
    The work closest to ours is~\cite{Guo2023CoTV:Learning}, that uses DRL to jointly control traffic lights and autonomous vehicles.
    Though many implementation details are similar, our approach is conceptually very different: While~\cite{Guo2023CoTV:Learning} assume autonomous vehicles with control residing on the vehicle side, we consider a system where the infrastructure controls both traffic light signaling and speed advice which is then sent to the vehicles.
    We expect most vehicles and infrastructure to be equipped with communication means in the near future.
    Automated vehicles, on the other hand, may take a lot longer to be widely deployed.
    This conceptual difference also influences implementation details.
    Most importantly, we learn individual traffic signaling and speed advice agents for each intersection, which has been shown to outperform other approaches in TLC~\cite{Ault2021ReinforcementControl}.
    In contrast,~\cite{Guo2023CoTV:Learning} learn a single autonomous driving policy that has to generalize over different intersections.

\section{Joint Control of Traffic Signaling and vehicle speed advice}\label{sec:methodology}
    In this Section, we describe our methodology of implementing the joint control of traffic signaling and vehicle speed advice.
    In particular, we explain the incremental adaptions we apply to the work of Ault, et. al.~\cite{Ault2021ReinforcementControl}, which from here on we will refer to as RESCO. 
    The central result of RESCO was that, evaluated on a broad set of traffic light control benchmark scenarios, the DRL algorithms that performed best were a set of independent learners (IDQN and IPPO), rather than highly specialized methods from self-proclaimed state-of-the-art methods.
    We take this as justification to start out with the environment formulation of the RESCO agent and successively augment state and action space to model the availability of detailed traffic information through V2X communication as well as the means to propose speed advice to individual drivers.
    We obtain three different agents that we extensively evaluate in \sec{sec:evaluation}.
    Among other things, these agents differ in their respective observation space.
    \tab{tab:observation_tl} summarizes the observation spaces of all agents.
    We also note the dimensionality of the individual parts of the observation space, where $N_P$ is the number of traffic light phase options of an intersection and $N_L$ is the number of afferent lanes.
    At runtime, the values are concatenated into a vector that is used as input to the policy NNs.
    During our implementation, we found several issues, that we regard as shortcomings of the original RESCO implementation or that are not in line with the goal of this work.
    We adapt for these issues and show in \sec{sec:evaluation} that we still approximately reproduce the results of RESCO.
    In particular, these adaptions are:\\
    \textbf{Short Lanes:} Real world scenarios in the RESCO benchmark contain several relatively short lanes as an artifact of the lane definition of the SUMO simulator.
    As traffic lights observe (state space) and optimize for (reward function) vehicles only on afferent lanes, this significantly reduces the sensing distance.
    As a countermeasure, we define all lanes shorter than 15 meters as short lanes and include their direct predecessors into the state space of traffic lights as well as including vehicles into the respective reward function.
    For the vehicle speed advice control described in \sec{sec:methodology}, short lanes are left uncontrolled.\\
    \textbf{Normalization of lane density:} Instead of using the exact number of approaching vehicles as input to our NN, we normalize this number by the maximum capacity of the lane, so that the density ranges from zero to one.\\
    \textbf{Sensing radius:} In the RESCO paper, a 200 meter sensing radius is assumed.
    As we consider V2X communication networks, that provide low latency communication over large distances, we remove this limitation.\\
    \textbf{Traffic light state:} RESCO does not include the time that a traffic light has been showing the same phase.
    This means that decisions cannot be based on time passed.
    We here include this time as well as a Boolean that indicates if the minimum phase time has been surpassed.\\
    \textbf{Reward Function:} RESCO uses reward functions according to the respective algorithm.
    We here use the negative time spent on afferent lanes of the intersection
     \begin{equation}
        r_i^{t+1} = - \sum_{\forall l \in L_i} \sum_{v \in V_l} \tau(v,l,t)\,,
    \end{equation}
    where $L_i$ is the set of afferent lanes of intersection i, $V_l$ is the set of vehicles that are currently on lane l, and $\tau(v,l,t)$ is the time that vehicle $v$ has spent on lane $l$ up to timepoint t.
    We chose this reward function over other commonly used ones, as it resulted in the best reduction of overall vehicle trip time in a comparative study we conducted (not reported here).
    
    Please note, that we strictly add information to the observation space.
    The resulting agent should therefore perform equally well or better than RESCO's IPPO implementation.
    We call the traffic light control agent, that results from the described adaptions, the "TLC" agent.
    It serves as a baseline to evaluate the benefits of equipping the system with V2X and vehicle speed advice.
    
    The first major adaption to the RESCO benchmark environments that we want to investigate, is the addition of detailed state information of individual vehicles that are made available through V2X communication means.
    Previous work demonstrated that the availability of detailed traffic state knowledge enables more efficient traffic light control~\cite{Busch2020OptimisedKnowledge}.
    However, in contrast to~\cite{Busch2020OptimisedKnowledge}, we here assume a baseline agent that includes live information about the density and average speed on individual lanes, which could be inferred approximately through inductive loops.
    The V2X-enabled agent is provided with detailed state information of the lane leader for every afferent lane of the controlled intersection.
    A lane leader is here defined as the vehicle that is closest to the intersection but is not part of the traffic light queue (so the vehicle is still moving).
    In particular, the additional information are the current distance from the next traffic light, the distance to the back of the traffic light queue, and the current speed of all lane leaders.
    We denote this agent the "TLC+V2X" agent.
    
    The final addition to RESCO is the implementation of vehicle speed advice.
    We consider two individual RL agents per intersection: one agent controlling traffic lights, as described above, and one agent controlling speed advice of individual vehicles on afferent roads.
    In particular the speed advice agent controls the speed advice of the lane leader of every afferent lane of the intersection (except for short lanes).
    This loosely corresponds to the case of speed advice being transmitted to individual vehicles via V2X communication. 
    We control the speed by incrementally adapting it once every 5 seconds of simulated time (same as for traffic light control).
    Each incremental adaption is chosen from a continuous interval $[-\Delta v_{max}, +\Delta v_{max}]$, where $\Delta v_{max}$ is set to ten percent of the speed limit of the respective lane.
    To realize these incremental updates, we added the current speed advice as an additional feature to the state space of the speed advice agent (other than that it uses the same state space as the TLC+V2X agent).
    In our experiments, controlling only lane leaders resulted in better performance than controlling all vehicles; and (relatively small) incremental updates, than absolute values.
    In addition, we hypothesize that these incremental adaptions with relatively small step sizes and relatively few updates would meet better acceptance by real drivers than large fluctuations on short timescales.
    Vehicles acceleration is controlled by SUMO's standard driver model~\cite{Treiber2000CongestedSimulations} that uses the vehicle speed advice as speed limit.
    SUMO's speed factors are preserved, which means that vehicles may drive slightly faster or slower than the speed limit/advice which results in additional randomness of the environment.
    We call this the "TLC+V2X+VSA" agent.

    \begin{table}[t]
        \centering
        \caption{%
        Traffic light observation space of the three agents compared in this paper (TLC, TLC+V2X, and TLC+V2X+VSA).
        [Square brackets] denote implementation differences in~\cite{Ault2021ReinforcementControl}.}
        \resizebox{\columnwidth}{!}{%
        \begin{tabular}{lc} 
            \toprule
            \textbf{Feature} & \textbf{Dimensions} \\ 
            \toprule
            \multicolumn{2}{c}{\emph{TLC agent [RESCO]}} \\
            {Current phase of the traffic light} & {$N_{P}$} \\
            {Lane density [number of vehicles on lane]} & {$N_L$} \\
            {Queue length per lane} &  {$N_L$} \\
            {Total waiting time of stopped vehicles per lane} & {$N_L$} \\
            {Average speed per lane} & {$N_L$} \\
            {Minimal green time passed [not included]} & {$1$} \\
            {Phase duration of the traffic light [not included]} & {$1$} \\
            \midrule 
            \multicolumn{2}{c}{\emph{Added through V2X (TLC+V2X agent)}} \\
            {Distance to traffic light of leading vehicle} & {$N_L$} \\
            {Distance to standing vehicle of leading vehicle}  & {$N_L$} \\
            {Speed of leading vehicle} & {$N_L$} \\
            \midrule 
            \multicolumn{2}{c}{\emph{Only TLC+V2X+VSA agent}} \\
            {Current speed limit per lane} & {$N_L$} \\
            \bottomrule
        \end{tabular}
        }
        \label{tab:observation_tl}
    \end{table}

\section{Experimental Setup}\label{sec:exp_setup}
    The experimental setup largely coincides with the one from RESCO.
    We therefore only briefly discuss it here.
    
    To test the efficacy of our method, we implement several benchmark scenarios in SUMO~\cite{Behrisch2011SUMOOverview}, which allows for the extraction of a detailed traffic state and the adaption of traffic lights and speed limits via the TraCI API.
    We run our experiments on eleven different scenarios, of which eight are taken from the RESCO benchmark.
    The RESCO scenarios consist of popular real-world benchmark scenarios from Ingolstadt (InTAS~\cite{Lobo2022InTASSUMO} -- a single intersection with 1715 vehicles approaching per hour, a corridor of seven intersections with 3030 vehicle per hour, and a patch of 21 intersections with 4280 vehicles per hour) and Cologne (Tapas~\cite{Varschen2006MikroskopischeZeitverwendungstagebuchern} -- a single intersection with 2015 vehicles per hour, a corridor of three intersections with 2856 vehicles per hour, and a patch of eight intersections with 2046 vehicles per hour).
    In addition, RESCO uses two synthetic scenarios of a four-by-four grid, with one encountering balanced traffic of 1473 vehicles per hour and the other one, highly frequented avenues on one axis and calmer roads on the other axis with 2484 vehicles per hour.
    All RESCO scenarios use predetermined vehicle spawn times and routes.
    Randomness of the trial runs stems only from random vehicle speed factors. 
    In addition to the RESCO scenarios, we implement a single intersection scenario for the detailed qualitative analysis of the the agent's behavior.
    The two perpendicular one-way streets consist of two lanes each.
    The traffic light therefore only has two different phase options.
    Vehicles may only go straight at the intersection and are created at the in-going roads following a binomial distribution.
    We simulate three different traffic demands: one low demand of approx. 70 vehicles per hour, one moderate traffic demand of approx. 500 vehicles per hour, and one high demand of approx. 2500 vehicles per hour.
    To switch phase, the traffic light goes through an amber phase that lasts two seconds.
    In addition, we enforce a minimal phase time of eight seconds.
    
    The performance metric we care about is average trip delay, which is the average trip time of vehicles minus the minimal possible trip time (of vehicles unconstrained by traffic or traffic lights).
    The reported values are obtained by running five trainings with different random seeds, computing mean and standard deviation over runs (not over episodes), and taking the minimum over episodes. 
    There might be an argument for reporting outcomes of multiple test runs of the best obtained model, when researching traffic control as an application rather than benchmarking DRL algorithms.
    However, we choose not to diverge from \cite{Ault2021ReinforcementControl}.
    In our experiments we also investigated average CO$_2$ emissions per trip.
    However, against our expectations, vehicle speed advice only resulted in minimal differences in CO$_2$ emissions in our experiments.
    Due to space constraints, we do not consider emissions in this publication.
    
    To train our IPPO agents, we use the Ray RLlib Python library~\cite{Liang2018RayLibrary}.
    For most hyperparameters, we use the provided default values.
    Exceptions are the learning rate, which we set to $10^{-5}$, the number of training episodes, which we set to 1400, and the DNN shape, which in our agents consists of four layers of 256 neurons each.

\section{Evaluation and Discussion}\label{sec:evaluation}
    \begin{table*}[t]
        \centering
        \caption{
        Mean and standard deviation of average trip delays in seconds for our 3 algorithms and IDQN and IPPO from \cite{Ault2021ReinforcementControl}.
        (Brackets) denote relative performance w.r.t. the left-hand result.
        Values marked with asterisk$^*$ were inferred from RESCO's training logs.
        As we were able to approximately, but not fully, reproduce the results of RESCO, we focus our comparison on our own algorithms.
        The joint optimization of speed advice and traffic signaling decreases trip delays in 8 out of 11 scenarios.}
        \resizebox{\textwidth}{!}{%
            \begin{tabular}{ll|ll|lll} 
                \toprule
                & & \multicolumn{2}{c}{Ault, et. al. \cite{Ault2021ReinforcementControl}} & \multicolumn{3}{c}{This work} \\
                & & IDQN & IPPO & TLC & TLC+V2X & TLC+V2X+VSA \\
                \midrule
                \multirow{3}{12mm}{Single\\intersec.} 
                    & Low dem. & n.a. & n.a. & $1.55 \hspace{1em} \pm 0.14 $  & $ 1.44 \hspace{1em} \pm 0.16 \hspace{.5em} \mathit{(-7.1\%)} $  & $ \mathbf{1.36 \hspace{1em} \pm 0.33} \hspace{.5em} \mathit{(-5.6\%)} $ \\
                    & Mod. dem. & n.a. & n.a. & $ 5.48 \hspace{1em} \pm 0.18 $ & $ 3.66 \hspace{1em} \pm 0.24 \hspace{.5em} \mathit{(-33.2\%)} $  & $ \mathbf{3.47 \hspace{1em} \pm 0.22} \hspace{.5em} \mathit{(-5.2\%)} $ \\
                    & High dem. & n.a. & n.a. & $ 9.95 \hspace{1em} \pm 0.38 $ & $ 10.05 \hspace{.5em} \pm 0.36 \hspace{.5em} \mathit{(+1.0\%)}$ & $ \mathbf{9.26 \hspace{1em} \pm 0.14} \hspace{.5em} \mathit{(-7.4\%)}$ \\
                    \cmidrule{2-7}
                \multirow{8}{12mm}{RESCO\\bench-\\mark} 
                    & Col. single & $26.05 \hspace{1em} \pm 0.57$ & $55.07 \hspace{.5em} \pm 11.83$ & $25.12 \hspace{.5em} \pm 0.64 \hspace{.5em} \mathit{(-3.5\%})$ & $\mathbf{23.07 \hspace{.5em} \pm 0.42} \hspace{.5em} \mathit{(-8.2\%)}$ & $24.35 \hspace{.5em} \pm 1.30\hspace{.5em} \mathit{(+5.5\%)}$ \\
                    & Col. corr. & $23.99 \hspace{1em} \pm 1.11$ & $22.13 \hspace{.5em} \pm 0.41$ & $22.40 \hspace{.5em} \pm 2.07 \hspace{.5em} \mathit{(+1.2\%)}$ & $22.00 \hspace{.5em} \pm 3.29 \hspace{.5em} \mathit{(-1.8\%)}$ & $\mathbf{20.99 \hspace{.5em} \pm 1.99 \hspace{.5em} \mathit{(-4.5\%)}}$ \\
                    & Col. reg. & $22.06 \hspace{1em} \pm 0.36$ & $21.49 \hspace{.5em} \pm 0.13$ & $20.85 \hspace{.5em} \pm 0.49 \hspace{.5em} \mathit{(-3.0\%)}$ & $19.98 \hspace{.5em} \pm 0.49 \hspace{.5em} \mathit{(-4.2\%)}$ & $\mathbf{19.48 \hspace{.5em} \pm 0.52 \hspace{.5em} \mathit{(-2.5\%)}}$ \\
                    & Ing. single & $21.48 \hspace{1em} \pm 0.56$ & $19.85 \hspace{.5em} \pm 0.21$ & $21.14 \hspace{.5em} \pm 0.59 \hspace{.5em} \mathit{(+1.5\%)}$ & $21.27 \hspace{.5em} \pm 0.46 \hspace{.5em} \mathit{(+0.6\%)}$ & $\mathbf{20.14 \hspace{.5em} \pm 0.34} \hspace{.5em} \mathit{(-5.3\%)}$ \\
                    & Ing. corr. & $31.19 \hspace{1em} \pm 0.97$ & $30.70 \hspace{.5em} \pm 0.98$ & $24.30 \hspace{.5em} \pm 0.59 \hspace{.5em} \mathit{(-20.8\%)}$ & $\mathbf{24.09 \hspace{.5em} \pm 0.38} \hspace{.5em} \mathit{(-0.8\%)}$ & $25.77 \hspace{.5em} \pm 0.36 \hspace{.5em} \mathit{(+7.8\%)}$ \\
                    & Ing. reg. & $59.64 \hspace{1em} \pm 2.13$ & $67.65 \hspace{.5em} \pm 19.49$ & $80.28 \hspace{.5em} \pm 7.08 \hspace{.5em} \mathit{(+34.6\%)}$ & $75.13 \hspace{.5em} \pm 10.89 \mathit{(-6.4\%)}$ & $\mathbf{59.84 \hspace{.5em} \pm 3.97} \hspace{.5em} \mathit{(-20.4\%)}$ \\
                    & Grid & $32.95 \hspace{1em} \pm 0.26^*$ & $44.00 \hspace{.5em} \pm 0.0^*$ & $56.98 \hspace{.5em} \pm 0.49 \hspace{0.5em} \mathit{(+72.9\%)}$ & $\mathbf{38.68 \hspace{.5em} \pm 0.78} \hspace{.5em} \mathit{(-32.1\%)}$ & $43.40 \hspace{.5em} \pm 0.60 \hspace{.5em} \mathit{(+12.2\%)}$ \\
                    & Avenues & $1168.32 \pm 194.45^*$ & $686.62 \pm 0.0^*$ & $794.86 \pm 98.00 \mathit{(+15.8\%)}$ & $645.41 \pm 31.74 \mathit{(-18.8\%)}$ & $\mathbf{594.99 \pm 21.79} \mathit{(-7.9\%)}$ \\
                \bottomrule
            \end{tabular}
        }
        \label{tab:quantitative_comparison}
    \end{table*}

    We split the evaluation into two parts: 
    First, we compare the performance of implemented agents in all implemented scenarios to analyze the quantitative benefit of vehicle speed advice.
    Second, we will analyze the single intersection scenario in depth to understand the qualitative behavior of the speed advice agent.

    As described in \sec{sec:exp_setup}, we compare our developed algorithms on the RESCO benchmark set as well as on a single intersection with three different traffic demands.
    \tab{tab:quantitative_comparison} shows the trip delay values for the IDQN and IPPO algorithm from~\cite{Ault2021ReinforcementControl} (that outperformed other investigated methods) and the results of this work.
    Results from RESCO are copied from \cite{Ault2021ReinforcementControl}; and for the single intersection scenarios, are not available.
    Values marked with an asterisk$^*$ were not reported in~\cite{Ault2021ReinforcementControl} but are taken from training logs in RESCO's Github repository.
    Numbers in (brackets) denote the relative performance of each agent w.r.t. the agent in the column to the left.
    As described in \sec{sec:methodology}, our TLC agent should be approximately equal to the IPPO agent from RESCO.
    As we do not observe the high instability of PPO that was reported in~\cite{Ault2021ReinforcementControl}, we compare our TLC agent against the better of RESCO's IDQN and IPPO agents for each scenario.
    The relative performance of our TLC agent therefore is w.r.t. the better one of RESCO's agents.
    Deviations range from 1.2~\% in Col. Corr up to 34.6~\% in Ing. Reg. (not counting values marked with $^*$, that we are unsure about).
    However, since it is often difficult to exactly reproduce findings from different implementations in a noisy domain like DRL, we still consider our TLC algorithm to approximately, but not fully, reproduce the results from~\cite{Ault2021ReinforcementControl}.
    Due to the strong deviations from the RESCO paper, our further analysis will focus on performance improvements among the algorithms from our own implementation.
    The introduction of detailed state knowledge via V2X results in improved trip delays in nine out of eleven scenarios.
    This is consistent with previous work~\cite{Busch2020OptimisedKnowledge}.
    Adding speed advice to the system further improves the performance in eight of the eleven scenarios.
    This shows that the use of speed advice in traffic systems can reduce trip delays of vehicles.
    As the TLC+V2X+VSA agent could theoretically learn to not use speed advice at all (and therefore reproduce the TLC+V2X agent), the increased trip delays in three of the eleven scenarios can only stem from sub-optimal learned policies.
    Further research therefore has to investigate better implementation details of the traffic control system.

\begin{figure}[t]
        \centering
        \includegraphics[width=\columnwidth]{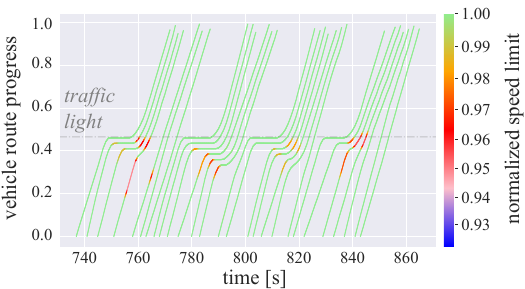}
        \caption{Route progress on the rightmost lane of the west-east route over time of multiple vehicles in the single intersection scenario with high demand.
        The position is normalized to the route length.
        Speed advice is color coded.
        Speed can be inferred from the gradients.
        The position of the traffic light is marked by the dashed line.
        This qualitative investigation shows that the speed advice agent lets vehicles close to a red traffic light slow down to smooth the velocity profile.}
        \label{fig:single_intersection_speed_advice}
    \end{figure}

    The atomic setting of one single intersection allows us to study the qualitative effects of implementing vehicle speed advice.
    \fig{fig:single_intersection_speed_advice} shows the position over time of multiple vehicles on the rightmost lane of the west-east route in the single intersection (high demand) scenario.
    The position is normalized to the length of the route.
    Each colored line represents a single vehicle traversing this route.
    The color indicates the speed advice given, normalized to the speed limit of the road.
    The dashed line marks the position of the traffic light.
    The speed advice agents slows down vehicles that are in relatively close proximity to the traffic light to smooth the velocity profile.
    Surprisingly, a relatively small adjustment of speed advice results in a relatively large improvement in trip delays.
    In fact, the maximum change in speed advice per decision, of ten percent of the road's speed limit, is rarely executed, justifying the choice of a small dynamic range.
    As an artifact of our implementation, which applies speed advice always to the first moving vehicle per lane, we can see speed advice decisions being passed on by vehicles that leave a lane, due to it being updated only every 5 seconds.
    Future work might experiment with different control frequencies as well as larger dynamic ranges of the speed advice.

\section{Conclusion and Outlook}
    As a first step towards jointly optimizing traffic light signaling and vehicle speed advice, we investigated the use of two independent PPO agents per intersection, one for each traffic control measure, to minimize average trip delay.
    We first approximately reproduced previous results on a set of benchmark scenarios~\cite{Ault2021ReinforcementControl}.
    Subsequently, we equipped the traffic infrastructure with a) more detailed traffic state knowledge and b) the capability to send speed advice to vehicles on afferent roads.
    Prior work already demonstrated that a more detailed traffic state facilitates better control in traffic lights~\cite{Busch2020OptimisedKnowledge}.
    Adding a learning agent to control vehicle speed advice, showed to further decrease average vehicle trip delays in eight of the eleven investigated scenarios.
    Analyzing the behavior of the speed advice agent, we found that this is obtained through slightly slowing vehicles close to the intersection to smooth out acceleration profiles.
    The conclusion we draw from this, is that the use of traffic speed advice, enabled through enhanced connectivity of vehicles and traffic infrastructure, has the potential to increase the efficiency of traffic systems and mitigate congestion. 
    Further we infer that DRL seems to be a fitting tool to learn the joint control of traffic signaling and vehicle speed advice.
    
    Though we obtained promising results using independent learners, future work needs to further optimize implementation details of the traffic control system.
    Furthermore, in theory, smoothing out velocity profiles should reduce CO$_2$ emissions.
    However, as our investigations in this regard were inconclusive, we did not report them here and left a thorough analysis to future research.

\bibliographystyle{IEEEtran}
\bibliography{references}
\end{document}